 \documentclass{colt2018} % Include author names

\title{Distributed Linear Model Clustering over Networks: A Tree-Based Fused-Lasso ADMM Approach}

\usepackage{inputenc} % allow utf-8 input
\usepackage[T1]{fontenc}    % use 8-bit T1 fonts
\usepackage{hyperref}       % hyperlinks
\usepackage{url}            % simple URL typesetting
\usepackage{booktabs}       % professional-quality tables
\usepackage{amsfonts}       % blackboard math symbols
\usepackage{nicefrac}       % compact symbols for 1/2, etc.
\usepackage{microtype}      % microtypography
\usepackage{times}
\usepackage{float}
\usepackage{algorithm}
\usepackage{makecell}
\usepackage{multirow}
\usepackage{hhline}
\usepackage{algorithmic}
\usepackage{amsmath,bm}
\usepackage{amssymb}
\usepackage[shortlabels]{enumitem}
\usepackage{mathrsfs}
\usepackage{graphicx}
\usepackage{color}
\usepackage{soul}
\usepackage{wrapfig}

\renewcommand{\b}{\mathbf{w}}

\newcommand{\CC}{\mathbf{C}}

\newcommand{\D}{\mathbf{D}}

\newcommand{\GG}{\mathbf{G}}

\newcommand{\I}{\mathbf{I}}

\renewcommand{\P}{\mathbf{P}}

\renewcommand{\u}{\mathbf{u}}

\renewcommand{\v}{\mathbf{v}}

\newcommand{\w}{\mathbf{w}}

\newcommand{\X}{\mathbf{X}}

\newcommand{\x}{\mathbf{x}}

\newcommand{\y}{\mathbf{y}}
\newcommand{\z}{\mathbf{z}}

\newcommand{\0}{\mathbf{0}}
\newcommand{\1}{\mathbf{1}}

\renewcommand{\L}{\mathbf{L}}
\renewcommand{\H}{\mathbf{H}}
\newcommand{\ppi}{\mbox{\boldmath{$\pi$}}}

\newcommand{\vDelta}{\boldsymbol \Delta}

\newtheorem{assumption}{Assumption}

\newtheorem{myTheo}{Theorem}

\newtheorem{myRemark}{Remark}

 \coltauthor{\Name{Xin Zhang} \Email{xinzhang@iastate.edu}\\
 \addr Department of Statistics, Iowa State University
 \AND
 \Name{Jia Liu} \Email{jialiu@iastate.edu}\\
 \addr Department of Computer Science, Iowa State University 
 \AND
 \Name{Zhengyuan Zhu} \Email{zhuz@iastate.edu}\\
 \addr Department of Statistics, Iowa State University 
}

\begin{document}

\maketitle

\begin{abstract}
%We propose an adaptive fused-lasso based coefficient subgroup approach for decentralized network systems.
%The major goal is to improve the model estimation efficiency by aggregating the neighbors' information as well as identifying the cluster membership for each node in the network. In particular, a tree-based $\ell_1$ penalty is proposed to reduce the computation and communication costs. 
%We show that our proposed estimator guarantees model selection consistency and asymptotic normality. 
%Also, we design a decentralized generalized alternating direction method of multiplier algorithm for solving the objective function in parallel and establish its linear convergence speed. 
%We conduct thorough numerical experiments to verify our theoretical results, which show that our approach outperforms existing works in terms of estimation accuracy, computation speed and communication costs.
In this work, we consider to improve the model estimation efficiency by aggregating the neighbors' information as well as identify the subgroup membership for each node in the network. A tree-based $l_1$ penalty is proposed to save the computation and communication cost. We design a decentralized generalized alternating direction method of multiplier algorithm for solving the objective function in parallel. The theoretical properties are derived to guarantee both the model consistency and the algorithm convergence. Thorough numerical experiments are also conducted to back up our theory, which also show that our approach outperforms in the aspects of the estimation accuracy, computation speed and communication cost.
\end{abstract}

\section{Introduction}\label{sec:intro}

In this paper, we consider a fundamental {\em distributed} linear model clustering problem over networks
%(sometimes also referred to as subgroup analysis in the literature):
(also sometimes referred to as subgroup analysis in the statistics literature): 
%and multi-task learning in the machine learning):
Suppose there are $K$ nodes in the network, each of which holds a dataset that is denoted as $D_k=\{(\x_{k,i}, y_{k,i})\}_{i=1}^{n},$ where $\x_{k,i} \in \mathbb{R}^{d}$ and $y_{k,i} \in \mathbb{R}$ ($k=1,\ldots,K$) represent the $i$-th covariate vector and response in the $k$-th dataset, respectively; and $n$ denotes the size of the dataset.
%The number of nodes (hence datasets) $K$ is fixed and 
For ease of exposition, the size of each dataset is assumed to be balanced (i.e., all nodes have $n$ samples)\footnote{Our algorithms and results in this paper can easily be extended to cases with datasets of unbalanced sizes.}. 
Hence, the total sample size in the network is $N=Kn$. 
%The datasets are scattered and stored in the nodes of the network. 
We assume that there exist $S$ underlying clusters of the nodes, and the data pair $(y,\x)$ from the $s$-th cluster follows a common linear model:
\begin{equation}\label{Eq: Model}
y = \b_{s}^\top \x+\varepsilon,
\end{equation}
where $\b_{s}=[\b_{s,1},\cdots,\b_{s,d}]^\top$ is a $d$-dimensional coefficient vector for the $s$-th cluster, the independent error $\varepsilon$ has a zero mean and a known variance $\sigma^2.$ 
The linear model in (\ref{Eq: Model}) varies across the underlying clusters, i.e., the datasets in the same cluster $s$ share the same coefficient $\b_s$ and vice versa.
Our goal is to identify the cluster membership of each node and their corresponding coefficient. 
However, due to communication limitation or privacy restrictions, one cannot merge these datasets to a single location. 
Thus, the main challenge of this problem is to perform clustering and estimate the coefficients of each cluster in the network in a {\em distributed} fashion.

The above problem naturally arises in many machine learning applications.
For example, a wireless sensor network is deployed in a large spatial domain to collect and learn the relationship between the soil temperature $y$ and air temperature $\x$ \cite{lee2015spatio}.
The domain can be divided into several subregions due to the landcover types, such as forest and grassland, and temperature relationships may vary geographically:
sensors in the same subregion may share the same regression relationship, and the coefficients vary across different subregions.
Similar scenarios could also emerge in other applications, such as meta-analysis on medical data \cite{tang2016fused}, federated learning on the speech analysis \cite{konevcny2015federated}, to name just a few.

%The above problem naturally arises in many machine learning applications.
%For example, to analyze weather patterns, a number of sensors are deployed in a large spatial domain to collect and estimate temperature data and related covariates. 
%Each sensor can only communicate with its nearby neighbors due to communication range constraints. 
%Due to spatial non-stationarity, the domain can be divided into several unknown subregions and the relationship between temperature and covariates may vary geographically: 
%sensors in the same subregion may share the same regression relationship, and the coefficients vary across different subregion.
%%The regions near each other could share an almost identical linear relationship, while two regions far away from each other may have data following drastically different linear models. 
%Similar scenarios could also emerge in many other applications, such as meta-analysis on medical data \cite{tang2016fused}, federated learning on the speech analysis \cite{konevcny2015federated}, to name just a few.

Unfortunately, distributively clustering nodes based on regression model over networks is challenging as it includes two non-trivial {\em inter-dependent} and {\em conflicting} subtasks: i) statistical estimator design and ii) distributed optimization under the proposed estimator.
%Generally speaking, when it comes to distributed algorithms, one prefers a design with a low computation complexity and communication costs.
%Unfortunately, to provide satisfactory statistical and optimization convergence guarantee, one requires more information exchange among the nodes.
In the literature, there exists tree-based centralized estimator designs that achieve strong statistical performance guarantee with $\Theta(K)$ computational complexity (e.g., \cite{tang2016fused,li2018spatial}, see Section~\ref{sec:related} for detailed discussions).
However, the tree-based penalty architectures make it difficult to design distributed optimization algorithms.
On the other hand, there exist efficient distributed algorithms for solving related clustering problems over networks (e.g., \cite{jiang2018consensus,wang2018distributed,hallac2015network}, see Section~\ref{sec:related} for details).
However, it is unclear whether they could provide statistical performance guarantees, such as the selection consistency and estimation normality.
Moreover, they all suffer $O(K^2)$ computational and communication costs.
In light of the limitations of these existing work, in this paper, we ask the following fundamental question:
{\em Could we develop a new distributed approach to achieve both strong statistical performance guarantees and $\Theta(K)$ computation and communication costs?
}
In other words, could we achieve the best of both worlds of the existing methods in the literature?

In this paper, we show that the answer to the above question is {\em affirmative}.
The main contribution of this paper is that, for the first time, we develop a new minimum spanning tree (MST) based fused-lasso approach for solving the network clustering problem.
Our approach enjoys oracle statistical performance and enables low-complexity distributed optimization algorithm design with {\em linear} convergence rate.
The main results of this paper are summarized as follows:
\begin{list}{\labelitemi}{\leftmargin=1em}
\itemsep -0pt

  \item {\em Low-Complexity Estimator Design:} We propose a new MST-based penalty function for the clustering problem with $\Theta(K)$ complexity. 
  Specifically, by comparing the coefficient similarities between the nodes, we construct a minimum spanning tree from the original network graph and only the edge in the tree are considered in the penalty function. 
  Under this approach, the terms in the penalty function is reduced to $K-1$ (hence $\Theta(K)$ as opposed to $O(K^2)$).
  \item {\em Statistical Performance Guarantee:} Based on the MST structure, we propose the use of adaptive lasso to penalize the linear model coefficient differences. 
  We show that our proposed estimator enjoys elegant oracle properties (cf. \cite{fan2001variable}), which means that our method can identify the nodes' cluster memberships almost surely (i.e., with probability one) as the size of datasets $n$ increases and the estimators achieve asymptotic normality.
  \item {\em Distributed Optimization Algorithm Design:} Due to the restrictions imposed by the tree-based estimator design, traditional gradient- or ADMM-type distributed methods cannot be applied to solve the objective function and find the nodes' cluster memberships distributively.
% fail to distributively solve the nodes' cluster identities.  
  In this paper, we develop a novel decentralized generalized ADMM algorithm for solving the tree-based fused-lasso problem. 
  Moreover, we show that our algorithm has a simple {\em node-based} structure that is easy to implement and also enjoys the {\em linear} convergence.
\end{list}

Collectively, our results in this paper contribute to the theories of low-complexity model inference/clustering over networks and distributed optimization.
%The rest of the paper is organized as follows: In Section \ref{Section: Model}, we will describe our model and algorithm in detail. The theoretical properties of our algorithm are also provided in Section \ref{Section: Model}. We conduct thorough numerical experiments in Section \ref{Section: Numerical} to validate our theoretical results. 
%The discussion and conclusion are provided in Section \ref{Section: Conclusion}. 
Due to space limitation, we relegate most of the proof details to supplementary material.

\section{Related work} \label{sec:related}

In the literature, many approaches have been developed to cluster the heterogeneous data, such as the mixture model methods \cite{hastie1996discriminant,shen2015inference,chaganty2013spectral}, the spectral clustering methods\cite{rohe2011spectral}, etc. 
However, most of the literature focuses on clustering the obeservation $y,$ rather than the relationship between $y$ and covariate $\x.$
The authors of \cite{ma2017concave,ma2018explore} are the first few to investigate the network clustering problem under the subgroup analysis framework.
%and the penalty function based methods (\cite{ ma2017concave,ma2018explore,li2018spatial,tang2016fused,fan2018}). 
%The penalty-based framework has been actively studied in the literature. 
Specifically, they considered the pairwise fusion penalty term for clustering the intercepts and the regression coefficients, respectively. 
%In contrast, our work focuses on discovering the clustering structure across multiple datasets with a tree-based penalty term.
In \cite{tang2016fused}, the authors proposed a fused-lasso method termed FLARCC to identify heterogeneity patterns of coefficients and to merge the homogeneous parameter clusters across multiple datasets in regression analysis with $\Theta(K)$ computational complexity. 
%By exploring the implicit ordering of the coefficients, FLARCC reduces the number of the pairwise penalty terms and facilitates computation without undermining the statistical efficiency. 
%However, the potential relationships of the datasets, such as the spatial-temporal structure or the network graph structure in this work, are not considered in FLARCC. 
However, FLARCC does not exploit any spatial network structure to further improve the performance.
The authors of \cite{li2018spatial} proposed a spatially clustered coefficient (SCC) regression method, which is based on a minimum spanning tree (MST) of the network graph to capture the spatial relationships among the nodes.
By contrast, in our work, we adopt the penalty function based framework to recover clusters identities by adding a penalty term $P_{\lambda}(\b_1,\cdots,\b_K)$ to the ordinary least square problem for (\ref{Eq: Model}). 
Unlike all the above methods that are implemented on single centralized machine, a key distinguishing feature of our work is that we need to conduct clustering in a {\em distributed} fashion.
As will be shown later, we improve the tree-based fusion penalty approach proposed in \cite{li2018spatial} to enhance the estimation efficiency as well as significantly reduce the computation and communication load for distributed algorithm design.

Our work also contributes to the theory of distributed optimization over networks, which have attracted a flurry of recent research (see, e.g., \cite{nedic2009distributed,yuan2016convergence,shi2014linear,eisen2017decentralized}). 
In the general framework of distributed optimization, all $K$ nodes in a connected network distributively and collaboratively solve an optimization problem in the form of: $\min_{\b} f(\b) \triangleq \sum_{i=1}^{K}f_i(\b)$,
where each $f_i$ is the objective function observable only to the $i$-th node and $\b$ is a global decision variable across all nodes. 
By introducing a local copy $\b_i,$ the above distributed optimization problem  can be reformulated in the following penalized version of the so-called {\em consensus} form: $\min_{\b_i,i=1,\cdots,K} \sum_{i=1}^{K}f_i(\b_i)+\frac{\lambda}{2}\sum_{i,j=1, i \neq j}^{K}\pi_{i,j}\|\b_i-\b_j\|_{2}^2$,
where $\pi_{i,j}$ is a weight parameter for penalizing the disagreement between the $i$-th and $j$-th nodes.
%Compared with their centralized counterparts, decentralized optimization algorithms could be more robust to attacks and system failures, while achieving the same convergence rate \cite{lian2017can}. 
%
Interestingly, in this work, opposite to traditional distributed algorithms that focus on the consensus problems stated above, we consider whether there exists {\em disagreements} among the true $\{\b_i\}_{i=1}^{K}:$ The nodes are to be classified to several clusters and the nodes in each cluster share the same $\b$.
We note that the authors of \cite{jiang2018consensus} and \cite{wang2018distributed} also focused on discovering the clustering patterns among the nodes with decentralized algorithms. 
However, they adopted a {\em pairwise} penalty function to obtain consensus of the inner-cluster weights, which can be reformulated as the well-known Laplacian penalty \cite{ando2007learning}.
A main limitation of the Laplacian penalty is that it cannot shrink the pairwise differences of the parameter estimates to zero (which is also verified in our simulations).

The most related work to ours is \cite{hallac2015network}, where the network lasso method was introduced. 
In the network lasso method in \cite{hallac2015network}, the authors adopted an $\ell_2$ penalty for each edge in the network graph. 
They also proposed a distributed alternating direction method of multipliers (ADMM) to solve the network lasso problem. 
Our work differs from \cite{hallac2015network} in the following key aspects: 
1) The number of the penalty terms in \cite{hallac2015network} depends on the number of edges in the network graph, which yields an $O(K^2)$ computation complexity and is {\em unscalable} for the large-sized networks. 
In this paper, we consider a {\em tree-based} penalty function, which contains exactly $K-1$ penalty terms; 
2) The penalty function in the network lasso method \cite{hallac2015network} adopted the $\ell_2$ norm for the $\b_i - \b_j$ difference, while we consider an {\em adaptive} $\ell_1$ norm for the vector difference, which enjoys elegant {\em oracle} prosperities (i.e., the selection consistency and the asymptotic normality); 
3) The algorithm in \cite{hallac2015network} is based on the classical ADMM algorithm with two constraints on each edge, while we propose a new generalized ADMM method with only {\em one} constraint on each edge, which significantly reduces the algorithm's implementation complexity; 
4) We rigorously prove the statistical consistency and algorithmic convergence of our proposed approach, both of which were not studied in \cite{hallac2015network}.

\section{Model and problem statement}\label{Section: Model}

Given a network $G=(V,E)$, where $V$ and $E$ represent the node and edge sets, respectively, our goal is to estimate the coefficients $\{\b_i\}_{i=1}^{K}$ and determine the cluster membership for each node. 
This problem can be formulated as minimizing the following loss function:
\begin{equation}\label{Eq: Obj}
L_{\text{Graph}}(\b)=\frac{1}{2}\sum_{i=1}^{K}\|\y_i-\X_i\b_i\|^2+\sum_{(v_i,v_j)\in E}P_{\lambda_N}(\b_{i}-\b_{j}),
\end{equation}
where $v_i\in V$ denotes the $i$-th node in the network; $\y_i=[y_{i,1},\cdots,y_{i,n}]^\top \in \mathbb{R}^{n}$ and $\X_i=[\x_{i,1},\cdots,\x_{i,n}]^\top \in \mathbb{R}^{n\times d}$ represent the reponses and design matrix at the $i$-th node, respectively; and $P_{\lambda_N}$ is a penalty function with tuning parameter $\lambda_N$. 
Note that the objective function in (\ref{Eq: Obj}) consists of two parts: the first part is an ordinary least square (OLS) problem for all the coefficients $\b \triangleq [\b_1^\top, \cdots, \b_K^\top]^\top \in \mathbb{R}^{Kn};$ 
the second term is a penalty term designed to shrink the difference of any two coefficient vectors if the corresponding nodes are connected.  
Note that the second term in (\ref{Eq: Obj}) depends on the network topology.
Thus, we make the following assumption that is necessary to guarantee that the problem is well-defined in terms of estimation accuracy:
%%%%%%%%%%%%%%%%%%%%%%%%%%%%%%%%%%%%%%%
\begin{assumption}\label{Assum:network}
Given a connected network $G=(V,E)$, for any node $v_i$ from a cluster with more than two members, there exists another node $v_j$ from the same cluster such that the edge $(v_i,v_j) \in E$.
\end{assumption}
Under Assumption~\ref{Assum:network}, each node is connected with its members if the cluster size is larger than one. 
Hence, by removing inter-cluster edges, i.e., identifying edges with {\em non-zero} coefficient difference, the original network graph can be reduced into $S$ subgraphs, which are the subgroup clusters.
For the objective function in (\ref{Eq: Obj}), several important remarks are in order:

\begin{myRemark}
{\em
The penalty terms in the objective function (\ref{Eq: Obj}) consist of all {\em pairwise} coefficient differences among all edges in the network graph. 
If the penalty function is chosen as $P_{\lambda_N}(\b_i,\b_j)=\lambda_N\|\b_i-\b_j\|_2$, then Eq.~(\ref{Eq: Obj}) has the same form as in the network lasso method \cite{hallac2015network}. 
The objective function (\ref{Eq: Obj}) can also be viewed as a variant of the method proposed in \cite{ma2017concave}, where the penalty terms are all pairwise differences of the nodes, and hence the total number of the penalty terms is exactly $(K-1)K/2$. 
Thanks to Assumption \ref{Assum:network}, we only need to consider the difference of end nodes of edges. Thus, the number of penalty terms can be reduced to exactly $|E|.$
However, the value of $|E|$ still implies that the number of penalty terms in (\ref{Eq: Obj}) could scale as $O(K^2)$ if the network is dense, which will in turn result in heavy computation and communication loads as the network size gets large. 
To address the problem, we will propose a simplified tree-based penalty function in Section~\ref{Section: MST}. 
%In Section~\ref{section: stat}, we will analyze the statistical properties for the proposed estimator. 
%A decentralized optimization algorithm will be proposed in Section \ref{section: Opt}.
}
\end{myRemark}

\section{Problem reformulation: a tree-based approach}\label{Section: MST}

As mentioned earlier and has been long noted in statistics (see, e.g., \cite{tang2016fused,li2018spatial}) and optimization (see, e.g., \cite{chow2016expander}) communities, directly including all edges in penalty terms will incur high computational and communication complexity. 
To reduce the redundant penalty terms, several strategies have been proposed, including the order method in \cite{ke2015homogeneity,tang2016fused} and the minimum spanning tree (MST) approach in \cite{li2018spatial}. 
Specifically, in \cite{ke2015homogeneity,tang2016fused}, the authors first determined the OLS estimation of the coefficients and then ordered the coefficients. 
They then presumed that similar coefficients will be neighbors with high probability and only regularization terms associated with the adjacent coefficients are considered. 
By contrast, in \cite{li2018spatial}, the authors used the {\em spatial} distance to constructed an MST, and the penalty terms in the tree are preserved. 
In essence, these two strategies are tree-based approaches, with the only difference being the definitions of distance measure for the tree: the first one uses model similarity, while the second one uses spatial distances.
In this paper, we propose a new tree-based approach, where the distance measure for the tree can be viewed as integrating the above two measures in some sense.
Yet, we will show that this new distance measure achieves surprising performance gains.

Specifically, we construct an MST as follows: First, {\em local} OLS estimators are determined in each node individually: $\hat{\b}_{i,OLS}=[\X_i^\top\X_i]^{-1}[\X_i^\top\y_i]$. 
Then, the weight for two nodes is defined based on their local model similarity and their connection relationship in the graph as follows:
\begin{align}\label{Eq: Weight}
\tilde{s}_{i,j}=
\begin{cases}
\|\hat{\b}_{i,OLS}-\hat{\b}_{j,OLS}\|, & \text{ if } (v_i,v_j) \in E, \\
\infty, & \text{ otherwise.}
\end{cases}
\end{align}
The weight $\tilde{s}$ in (\ref{Eq: Weight}) contains two important pieces of information: one is the network topology, which is characterized by spatial distances (e.g., in a sensor network, the nodes can only be connected within a certain communication range); the other is the local model similarity, which implies the likelihood of two nodes being in the same cluster. 
Based on (\ref{Eq: Weight}), an MST can be constructed so that only penalty terms associated with the MST are considered in the objective function:
\begin{equation} \label{eqn_MSTs}
L_{\text{MST}_s}(\b)=\frac{1}{2}\sum_{i=1}^{K}\|\y_i-\X_i\b_i\|^2+ \!\!\!\! \sum_{(v_i,v_j)\in \text{MST}_s} \!\!\!\! P_{\lambda}(\b_{i}-\b_{j}),
\end{equation}
where the notation $\text{MST}_s$ signifies that the MST is based on the model \textbf{s}imilarity.
Note that the estimation efficiency and clustering accuracy significantly depend on the penalty function.
% (the tree structure in our case). 
The following lemma guarantees that the nodes in the same cluster are connected in the $\text{MST}_s$ based on the weight defined in (\ref{Eq: Weight}) (see Section~2.1 in supplementary material for proof details).
%%%%%%%%%%%%%%%%%%%%%%%%%%%%%%%%%%%%%%%%%%%%%%%%%%%%%%%%%
\begin{lemma}\label{Lemma: MST}
Under Assumption \ref{Assum:network}, given an $\text{MST}_{s}$ based on the weights defined in (\ref{Eq: Weight}), as the local sample size $n \rightarrow \infty,$ then with probability $1,$ for any node $v_i$ in a cluster $s$ with more than two members, there exists a node $v_j$ from the same cluster such that the edge $(v_i,v_j)$ is in the $\text{MST}_{s}$.
\end{lemma}
%Note that there are total $K-1$ edges in the $\text{MST}_{s}$, where $K$ is the node number. 
With Lemma \ref{Lemma: MST}, the number of inter-cluster edges is $S-1$. 
Thus, the $\text{MST}_{s}$ is a connected graph with the {\em smallest} possible number of inter-cluster edges. 
Also, the $\text{MST}_{s}$ can be separated into $S$ clusters by identifying these inter-cluster edges. 
We note that there exist distributed methods to find the MST$_s$ (e.g., the GHS algorithm \cite{gallager1983distributed}) and their implementation details are beyond the scope of this paper.

\section{Statistical model: an adaptive fused-lasso based approach}\label{section: stat}

For convenience, we use $[\v]_{p}$ to denote the $p$-th element of vector $\v$.
Based on the MST constructed in Section \ref{Section: MST}, we specialize the loss function in (\ref{eqn_MSTs}) by adopting the following adaptive lasso penalty: 
\begin{align}\label{Eq: MST_obj} 
L_{\text{MST}_s}(\b)=\frac{1}{2}\sum_{i=1}^{K}\|\y_i-\X_i\b_i\|^2+\frac{\lambda_N}{2}\sum_{i=1}^{K}\sum_{j\in \mathcal{N}_i}\sum_{p=1}^{d}[\hat{\ppi}_{i,j} ]_p \big|[\b_{i}]_p - [\b_{j}]_p \big|,
\end{align}
where $\mathcal{N}_i$ represents the set of the neighboring nodes of node $i$ in the MST$_s$, $\hat{\ppi}_{i,j} \in \mathbb{R}^{d}$ is an adaptive weight vector defined as $[\hat{\ppi}_{i,j}]_p=1/ \big|[\hat{\b}_{i,OLS}]_p-[\hat{\b}_{j,OLS}]_p \big|^\gamma$ for some constant $\gamma>0.$ Therefore, our proposed estimator is $\hat{\b}_{\text{MST}_s}=\arg\min_{\w} L_{\text{MST}_s}(\b).$

\begin{myRemark}
{\em
Here, our use of an adaptive lasso penalty is motivated by: 1) Adaptive lasso is known to be an oracle procedure for related variable selection problems in statistics \cite{fan2001variable}; 
2) With an adaptive lasso penalty, the objective function in (\ref{Eq: MST_obj}) is strongly convex as long as the design matrix $\X$ is of full row rank. This implies that the minimum of (\ref{Eq: MST_obj}) is unique. 
In \cite{ma2017concave,ma2018explore}, similar clustering methods were proposed based on the minimax concave penalty (MCP) and the smoothly clipped absolute deviations (SCAD) penalty, both of which are concave penalties and have been shown to be statistical efficient. 
However, from optimization perspective, concave penalties will render the objective function non-convex, which in turn lead to intractable algorithm design. 
In \cite{li2018spatial}, lasso penalty was also adopted, but there is no proof for the oracle prosperities of their estimator. 
}
\end{myRemark}

For more compact notation in the subsequent analysis, we rewrite the objective function (\ref{Eq: MST_obj}) in the following matrix form:
\begin{align}\label{Eq: MST_obj_matrix}
L_{\text{MST}_s}(\b)=\frac{1}{2}\|\y-\X\b\|^2+\lambda_N \sum_{p=1}^{d(K-1)} [\hat{\ppi}]_{p}\cdot[(\H\otimes \I_K) \b]_{p}|
\end{align}
where $\y=[\y_1^\top,\cdots,\y_K^\top]^\top,$ $\X=\text{diag}(\X_1,\cdots,\X_K)^\top$ and $\b=[\b_1^\top,\cdots,\b_K^\top]^\top$ are the response vector, the design matrix, and coefficient vector, respectively; and $\otimes$ denotes the Kronecker product. 
In (\ref{Eq: MST_obj_matrix}), $\H$ is the incident matrix of the $\text{MST}_s$, which is row full rank and each entry in $\H$ defined as:
%with each row presents a edge in the tree and each column presents one node, 
\begin{align}
[\H]_{l,i}=
\begin{cases}
1, & \text{ if } i=s(l), \\
-1, & \text{ if } i = e(l), \\
0, & \text{ otherwise},
\end{cases}
\end{align}
where $s(l)$ and $e(l)$ denote the starting and ending node indices of edge $l$ in the MST$_s$, respectively, with $s(l) < e(l)$.
In (\ref{Eq: MST_obj_matrix}), $[\hat{\ppi}]_{p} \triangleq1/[\underline{\H}\cdot \b_{OLS}]_{p}^\gamma$, where $\underline{\H} \triangleq \H \otimes \I_K$ and $\b_{OLS}=[\b_{1,OLS}^\top,\cdots,\b_{K,OLS}^\top]^\top$ is the vector form of the OLS estimations.
%$[\cdot]_p$ means the $p$th element in a vector.
Note that adding one more row to $\H$, we can form a square and full rank matrix:
$\tilde{\H}=\begin{bmatrix}
   \H\\
   \frac{1}{\sqrt{K}}\1^\top
\end{bmatrix}$ \cite{li2018spatial},
and the objective function (\ref{Eq: MST_obj_matrix}) can be equivalent rewritten as:
\begin{align} \label{eqn_obj_w}
L_{\text{MST}_s}(\b)=\frac{1}{2}\|\y-\X\b\|^2+\lambda_N \sum_{p=1}^{dK} [\hat{\ppi}]_{p}\cdot|[\tilde{\underline{\H}} \b]_{p}|,
\end{align}
%where the weight $\hat{\ppi}$ is defined as:
%%%%%%%%%%%%%%%%%%%%%%%%%%%%%%%%%%%%%%%%
%\begin{align}
%[\hat{\ppi}]_{p}=
%\begin{cases}
%1/[\tilde{\underline{\H}} \b_{OLS}]_{p}^{\gamma}, & \text{if }~p~\text{mod}~K\neq 0\\
%0, & \text{otherwise.}
%\end{cases}
%\end{align}
where $\tilde{\underline{\H}} \triangleq \tilde{\H}\otimes \I_K$ is a full rank square matrix. Define $\vDelta = \tilde{\underline{\H}} \b$ as the difference of the connected nodes' weights.
It then follows that the objective function in (\ref{eqn_obj_w}) can be rewritten in terms of $\Delta$ as: 
\begin{align}
L_{\text{MST}_s}(\vDelta)=\frac{1}{2}\|\y-\X\tilde{\underline{\H}}^{-1}\vDelta\|^2+\lambda_N \sum_{p=1}^{dK} [\hat{\ppi}]_{p}\cdot |[\vDelta]_{p}|.
\end{align}
Our estimator then becomes: $\widehat{\vDelta}_{\text{MST}_s} = \arg\min_{\vDelta} L_{\text{MST}_s}(\vDelta).$
%We note that similar reformulation techniques have also been used in \cite{tang2016fused,li2018spatial}. 
Since there is a one-to-one transformation between $\hat{\b}_{\text{MST}_s}$ and $\widehat{\vDelta}_{\text{MST}_s}$ (i.e., $\widehat{\vDelta}_{\text{MST}_s} = \tilde{\underline{\H}} \widehat{\b}_{\text{MST}_s}$), 
we can instead focus on the theoretical prosperities of $\widehat{\boldsymbol{\Delta}}_{\text{MST}_s}$.
Denote the true coefficients as $\b_*=[\b_{1,*}^\top,\cdots,\b_{K,*}^\top]^\top$, and $\vDelta_* = \tilde{\underline{\H}} \b_*.$ 
Note that if the two connected nodes are from the same cluster, the corresponding elements in $\vDelta_*$ are zero. 
We denote the set of non-zero elements in $\vDelta_*$ as $\mathcal{A}_*.$ 
Similarly, the set of non-zero elements in $\widehat{\vDelta}_{\text{MST}_s}$ is denoted as $\hat{\mathcal{A}}_{N}$.
To prove the oracle properties of $\widehat{\vDelta}_{\text{MST}_s}$, we need the following assumptions for the linear model in (\ref{Eq: Model}):
\begin{assumption}\label{Assum: model}
For the linear model in (\ref{Eq: Model}):
i) the errors are i.i.d. random variables with zero mean and variance $\sigma^2$;
ii) $\frac{1}{N}(\X\tilde{\underline{\H}}^{-1})^{\top}\X\tilde{\underline{\H}}^{-1}\xrightarrow{p} \CC$ for some positive definite matrix $\CC$ as $N \rightarrow \infty$.
\end{assumption}
%\begin{myRemark}
We note that in the second condition in Assumption~\ref{Assum: model}, since $\tilde{\underline{\H}}$ is a full rank square matrix, $\CC$ is positive definite if $\X$ is full column rank.
Now, we state the oracle properties of $\widehat{\vDelta}_{\text{MST}_s}$ as follows:
%\end{myRemark}
\begin{myTheo}\label{Theo: consistency}
Suppose that $\lambda_N/\sqrt{N}\rightarrow 0$ and $\lambda_N N^{(\gamma-1)/2}\rightarrow \infty.$ Under Assumptions~\ref{Assum:network} and \ref{Assum: model}, the estimator $\widehat{\vDelta}_{\mathrm{MST}_s}$ satisfies the following two oracle properties:
i) (Selection Consistency) $\lim_{n\rightarrow \infty} \mathbb{P}(\hat{\mathcal{A}}_{N}=\mathcal{A}_*)=1$;
and ii) (Asymptotic Normality) $\sqrt{N}([\widehat{\vDelta}_{\mathrm{MST}_s}]_{\mathcal{A}_*}-[\vDelta_*]_{\mathcal{A}_*}) \xrightarrow{d} \mathcal{N}(0,\sigma^2\CC_{\mathcal{A}_*}^{-1})$ as $N \rightarrow \infty,$ where $\CC_{\mathcal{A}_*}^{-1}$ is the submatrix of  corresponding to set $\mathcal{A}_*$.
\end{myTheo}
The proof of Theorem~\ref{Theo: consistency} is relegated to supplementary material due to space limitation.
Based on Theorem \ref{Theo: consistency}, the asymptotic normality for $\hat{\b}_{\text{MST}_s}$ can be derived by a simple linear transformation.

\section{Optimization algorithm: an ADMM based distributed approach}\label{section: Opt}

In this section, we will design a distributed algorithm for minimizing (\ref{Eq: MST_obj}).
Due to the penalty structure in (\ref{Eq: MST_obj}), one natural idea is to use the popular ADMM method \cite{Boyd11:ADMM_Book}, which has been shown to be particularly suited for solving lasso related problems (e.g., \cite{ma2017concave,ma2018explore,wahlberg2012admm,zhu2017augmented}). 
%{\color{red} However, to make the algorithm feasible for the network system, we adopt the generalized ADMM proposed in \cite{deng2016global}.}
However, in what follows, we will first illustrate why it is challenging to use a regular ADMM approach to solve the MST$_s$-based fused-lasso clustering problem over networks in a distributed fashion.
As a result, it is highly non-trivial to design a new ADMM-based algorithm by exploiting special problem structure in the MST$_s$ regularizer.
To this end, we first note that the penalty term in (\ref{Eq: MST_obj}) can be written as:
%%%%%%%%%%%%%%%%%%%%%%%%%%%%%%%%%%%%%%%
\begin{align} \label{eqn_penalty_reform}
\frac{1}{2}\sum_{i=1}^{K}\sum_{j\in \mathcal{N}_i}\sum_{p=1}^{d}[\hat{\ppi}_{i,j}]_p \big| [\b_{i}]_p-[\b_{j}]_p \big|=\sum_{e_l\in\text{MST}_s}\sum_{p=1}^{d}[\hat{\ppi}_{l}]_p \big|[\b_{s(l)}]_p - [\b_{e(l)}]_p \big|,
\end{align}
where $e_l$ represents the $l$-th edge in MST$_s$.
In (\ref{eqn_penalty_reform}), $s(l)$ and $e(l)$ denote the starting and ending node indices of edge $l$, respectively, with $s(l) < e(l)$;
and $\hat{\ppi}_{l}=\hat{\ppi}_{s(l),e(l)}$ is the corresponding adaptive weight vector for the $l$-th edge. 
With the same notation as in Section \ref{section: stat}, the weight difference at edge $l$ is $\vDelta_l=\b_{s(l)}-\b_{e(l)}$ %with $i<j$ 
and $\vDelta =[\Delta_1^\top,\cdots,\Delta_{K-1}^\top]^\top= \underline{\H}\b$.
Note that there are $K-1$ edges in the MST$_s$. 
Thus, the problem of minimizing the loss function in (\ref{Eq: MST_obj}) can be reformulated as: 
\begin{align}\label{Eq: Constrain_problem}
\text{Minimize } & \,\, \frac{1}{2}\sum_{i=1}^{K}\|\y_i-\X_i\b_i\|^2+\lambda_N\sum_{l=1}^{K-1}\sum_{p=1}^{d}[\hat{\ppi}_{l}]_p \big|[\vDelta_{l}]_p \big|\\
\text{subject to } & \,\, \vDelta =\underline{\H}\b. \notag
\end{align}
Then, we can construct an augmented Lagrangian with penalty parameter $\tau>0$ for (\ref{Eq: Constrain_problem}) as follows:
%%%%%%%%%%%%%%%%%%%%%%%%%%%%%%%%%%%%%%%
\begin{align}\label{Eq: full_lag}
\!\!\!\!\!\! L_{\tau}(\b,\vDelta,\z) \!=\! \frac{1}{2}\sum_{i=1}^{K}\|\y_i \!-\! \X_i\b_i\|^2 \!+\!\lambda_N \!\! \sum_{l=1}^{K-1}\sum_{p=1}^{d}[\hat{\ppi}_{l}]_p \big|[\vDelta_{l}]_p \big| \!-\! \langle \z,\underline{\H}\b \!-\! \vDelta\rangle \!+\! \frac{\tau}{2}\|\underline{\H}\b \!-\! \vDelta\|^2\!, \!\!\!\!\!\!\!
\end{align}
where $\z \in \mathbb{R}^{d(K-1)}$ is the vector of dual variables corresponding to the $K-1$ edges.
In what follows, we derive the updating rules for $(\b^{t+1},\vDelta^{t+1},\z^{t+1})$.
First, given the primal and dual pair $\b^{t}, \z^{t},$ 
for the $l$-th edge with end nodes $s(l)$ and $e(l)$, to determine the weight difference $\vDelta^{t+1}$, we need to solve the subproblem $\vDelta^{t+1}
= \arg\min_{\vDelta} L_\tau(\b^{t},\vDelta,\z^{t}),$ 
and hence for the $l$-th edge with end nodes $s(l)$ and $e(l)$, it follows that (see Section~1 in supplementary material for derivation details):
\begin{align}\label{Eq: updating_1}
\vDelta^{t+1}_l&
=S_{\lambda_N \hat{\ppi}_{l}/\tau}\Big(\b_{s(l)}^{t}-\b_{e(l)}^{t}-\frac{1}{\tau}\z_{l}^{t}\Big), %, ~\text{if} ~i<j
\end{align}
where $S_{\lambda_N \hat{\ppi}_{l}/\tau}$ is the coordinate-wise soft-thresholding operator with $[\lambda_N \hat{\ppi}_{l}/\tau]_p=\lambda_N [\hat{\ppi}_{l}]_p/\tau.$
Next, we derive the updating rule for $\b^{t+1}$. 
With the classical ADMM, it can be shown that (see Section~1 in supplementary material for derivation details):
\begin{align}\label{Eq: classical_admm}
\b^{t+1}
& 
=\arg\min_{\b} L_\tau(\b,\vDelta^{t+1},\z^{t})\notag\\ 
&
=[\X^\top\X+\tau \L\otimes \I_d]^{-1}[\X^\top\y + \underline{\H}^\top(\tau\vDelta^{t+1}+\z^t)].
\end{align}
Unfortunately, the matrix inverse in (\ref{Eq: classical_admm})  {\em cannot} be computed in a distributed fashion due to the coupled structure of the Laplacian matrix $\L.$ 
Here, we show that the generalized ADMM studied in \cite{deng2016global} can be leveraged to derive an updating rule for $\w^{t+1}$, which can be implemented in a parallel fashion. 
To this end, instead of directly solving the subproblem $\b^{t+1}=\arg\min_{\b} L_\tau(\b,\vDelta^{t+1},\z^{t})$, we add a quadratic term $\frac{1}{2}(\b-\b^t)^\top \P (\b-\b^t)$ in the subproblem ($\P$ is positive semidefinite):
\begin{align*}
\b^{t+1}
&
= \arg\min_{\b} L_\tau(\b,\vDelta^{t+1},\z^{t})+\frac{1}{2}(\b-\b^t)^\top \P (\b-\b^t)\\
&
=[\X^\top\X+\tau\underline{\H}^\top\underline{\H}+\P]^{-1}[\X^\top\y + \underline{\H}^\top(\tau\vDelta^{t+1}+\z^t)+\P\b^t].
\end{align*}
Now, the {\em key step} is to recognize that we can choose the matrix $\P= -\tau \underline{\H}^\top\underline{\H} + \D = -\tau \L\otimes \I_d + \D,$ where $\D= \text{diag}(D_1,\cdots,D_K)\otimes \I_d$ with positive scalars $D_i$ for node $i$ and $\L=\H^\top \H$ is the Laplacian matrix for the MST$_s$.
It follows that $\b^{t+1}=[\X^\top\X+\D]^{-1}[\X^\top\y + \underline{\H}^\top(\tau\vDelta^{t+1}+\z^t)+\P\b^t]$.
Plugging in $\P=-\tau \L\otimes \I_d + \D$, we have the following local weight update: 
\begin{equation}\label{Eq: updating_2}
\b_{i}^{t+1}=[\X_i^\top\X_i+D_i\I_d]^{-1} \Big[\X_i^\top\y_i + \sum_{v_i\in e_l} [\H]_{li}(\tau\vDelta_l^{t+1}+\z_l^t)+(D_i-\tau \text{deg}(i)\b_i^t+\tau\sum_{j\in \mathcal{N}_i}\b_j^t \Big],
\end{equation}
where $v_i\in e_l$ means node $v_i$ is an end node of edge $e_l,$ $\text{deg}(i)$ is the degree of the node $v_i$ (i.e., $\text{deg}(i) = |\mathcal{N}_{i}|$. 
Thus, the updating of $\b_i^{t+1}$ only requires the local and connected neighbor's information, which facilitates {\em distributed} implementation. 
Also, matrix $\D$ plays an important role on the algorithm convergence.
Recall that $\P=\D -\tau L\otimes \I_d  =[\text{diag}(D_1,\cdots,D_K)-\tau L]\otimes \I_d.$ 
To guarantee $\P\succ 0,$ based on the Gershgorin circle theorem, we can choose $\D$ as $D_i> 2\text{deg}(i)$.
Lastly, the dual variables $\z^{t+1}$ can be updated as $\z^{t+1}=\z^{t}-\tau(\underline{\H}\b^{t+1}-\Delta^{t+1})$,
and hence for the $l$-th edge, the corresponding dual update is (see Section~1 in supplementary material for details): 
\begin{equation}\label{Eq: updating_3}
\z_l^{t+1}=\z_l^{t}-\tau \Big(\b_{s(l)}^{t+1}-\b_{e(l)}^{t+1}-\vDelta_l^{t+1} \Big).
\end{equation}

Note, however, that the updating rules (\ref{Eq: updating_1}) and (\ref{Eq: updating_3}) are edge-based while (\ref{Eq: updating_2}) is node-based. 
To make the updateing rules consistent, we define several additional notations: 
At node $s(l)$, we let $\vDelta_{s(l)}^{t}=\vDelta_l^{t}$ and $\z_{s(l)}^{t} = \z_l^{t}$;
At node $e(l)$, we let $\vDelta_{e(l)}^{t}=-\vDelta_l^{t}$ and $\z_{e(l)}^{t}=-\z_l^{t}$. 
With simple derivations, it can be verified that if $\vDelta_{s(l)}^{t} = -\vDelta_{e(l)}^{t} = \vDelta_l^{t}$ and $\z_{s(l)}^{t}=-\z_{e(l)}^{t}=\z_l^{t}$ are satisfied in iteration $t$, then in iteration $t+1$, $\vDelta_{s(l)}^{t+1}=-\vDelta_{e(l)}^{t+1}=\vDelta_l^{t+1}$ and $\z_{s(l)}^{t+1}=-\z_{e(l)}^{t+1}=\z_l^{t+1}$ still hold based on the following node-based updating rules: $\forall i \in \{s(l), e(l)\}$ and $j = \{s(l), e(l)\}/\{i\},$
\begin{align}\label{Eq: updating_node}
\left\{
\begin{aligned}
&\vDelta^{t+1}_{i}=S_{\lambda_N \hat{\ppi}_{l}/\tau}\Big(\b_{i}^{t}-\b_{j}^{t}-\frac{1}{\tau}\z_{i}^{t}\Big), \\
&\b_{i}^{t+1}\!=\![\X_i^\top\X_i \!+\! D_i\I_d]^{-1} \Big[\X_i^\top\y_i \!+\! \sum_{v_i\in e_l} (\tau\vDelta_{i}^{t+1} \!+\!\z_{i}^t) \!+\! (D_i \!-\! \tau \text{deg}_i)\b_i^t \!+\! \tau\sum_{j\in \mathcal{N}_i}\b_j^t \Big], \\
&\z_{i}^{t+1}=\z_{i}^{t}-\tau \Big(\b_{i}^{t+1}-\b_{j}^{t+1}-\Delta_{i}^{t+1} \Big).
\end{aligned}
\right.
\end{align}
Thus, we can set $\vDelta_{s(l)}^0 = - \vDelta_{e(l)}^0 =\b_s(l)^0 - \b_e(l)^0$ and $\z_{s(l)} = \z_{e(l)} =\0,$ $\forall l$, which satisfy the above conditions. 
Note that the updating rule for $\b_{e(l)}^{t+1}$ has the same structure as $\b_{s(l)}^{t+1}$ because $[\H]_{l,e(l)}=-1$ and $\z_{e(l)}^{t}=-\z_{l}^{t},$ $\vDelta_{e(l)}^{t}=-\vDelta_{l}^{t}$. 
Our method is summarized in Algorithm \ref{Alg: ADMM}. 
The outputs of the algorithm are the estimated coefficient $\hat{\b}$ and the coefficient difference $\widehat{\Delta}$. 
Whether two nodes are in the same cluster can be determined by checking $\widehat{\vDelta}:$ $\widehat{\vDelta}_{s(l)}=\widehat{\vDelta}_{e(l)}=\0$ if $s(l)$ and $e(l)$ are in the same cluster.
The following theorem guarantees the convergence speed of Algorithm~\ref{Alg: ADMM}.
\begin{myTheo}\label{Theo: convergence}
Denote the KKT point for the objective function (\ref{Eq: Constrain_problem}) as $\u_*=(\b_*^\top,\vDelta_*^\top,\z_*^\top)^\top$. 
With a proper selection of $\D$ such that $\P \succ 0,$ the iterates $\{\u^{t}\}_{t=1}^{\infty}$ converge to $\u_*$ in the sense of $\GG$-norm: $\|\u^{t}-\u_*\|_{\GG} \rightarrow 0,$ where $\|\cdot\|_{\GG}$ represents the semi-norm $\|\x\|^2_{\GG} \triangleq \x^\top \GG \x $, where $\GG$ is defined as:
\begin{align*}
\GG \triangleq \begin{bmatrix}
   \D & \0 & \0 \\
    \0  & \0 & \0 \\
    \0  & \0 & \frac{1}{\tau}\I_{d(K-1)} \\
\end{bmatrix}, 
\end{align*}
Further, the convergence rate is linear, i.e., $\exists~ \delta>0,$ such that $\|\u^{t+1}-\u_*\|^2_{\GG} \leq (1+\delta)^{-1}\|\u^{t}-\u_*\|^2_{\GG}$.
\end{myTheo}

\begin{algorithm}[t!]
\caption{Decentralized Generalized ADMM for Minimizing $L_{\text{MST}_s}$ in (\ref{Eq: MST_obj}).}\label{Alg: ADMM}
\begin{algorithmic} [1]
\REQUIRE Data $\{\X_i, \y_i\}_{i=1}^{K},$ tuning parameter $\lambda_N;$
\ENSURE $\b^{T}$ and $\vDelta^{T};$
\STATE Each node finds the local OLS estimation and sets $\b_i^0=\b_{i,OLS}$;
\STATE Each node sends $\b_{i}^0$ to its neighboring nodes in the network and calculates the weight (\ref{Eq: Weight});
\STATE The network constructs an MST$_s$ based on $\w_{i}^{0}$;
\STATE The nodes $s(l)$ and $e(l)$ of edge $l$ set $\z_{s(l)}^0=\z_{e(l)}^0=\0$ and $\vDelta_{s(l)}^0=-\vDelta_{e(l)}^0=\b_{s(l)}^0-\b_{e(l)}^0$. %with $e_l=(v_i,v_j);$
\WHILE{not converged}
    \STATE Each node sends its current $\b_i^t$ to its neighboring nodes in the MST$_s$;
    \STATE Each node updates the primal and dual variables using the rules in (\ref{Eq: updating_node}).
\ENDWHILE
\end{algorithmic}
\end{algorithm}

\section{Numerical Results}\label{Section: Numerical}

\begin{wrapfigure}{r}{0.5\textwidth}
\centering
\vspace{-.15in}
\begin{tabular}{@{}cc@{}}
\includegraphics[width=.47\linewidth]{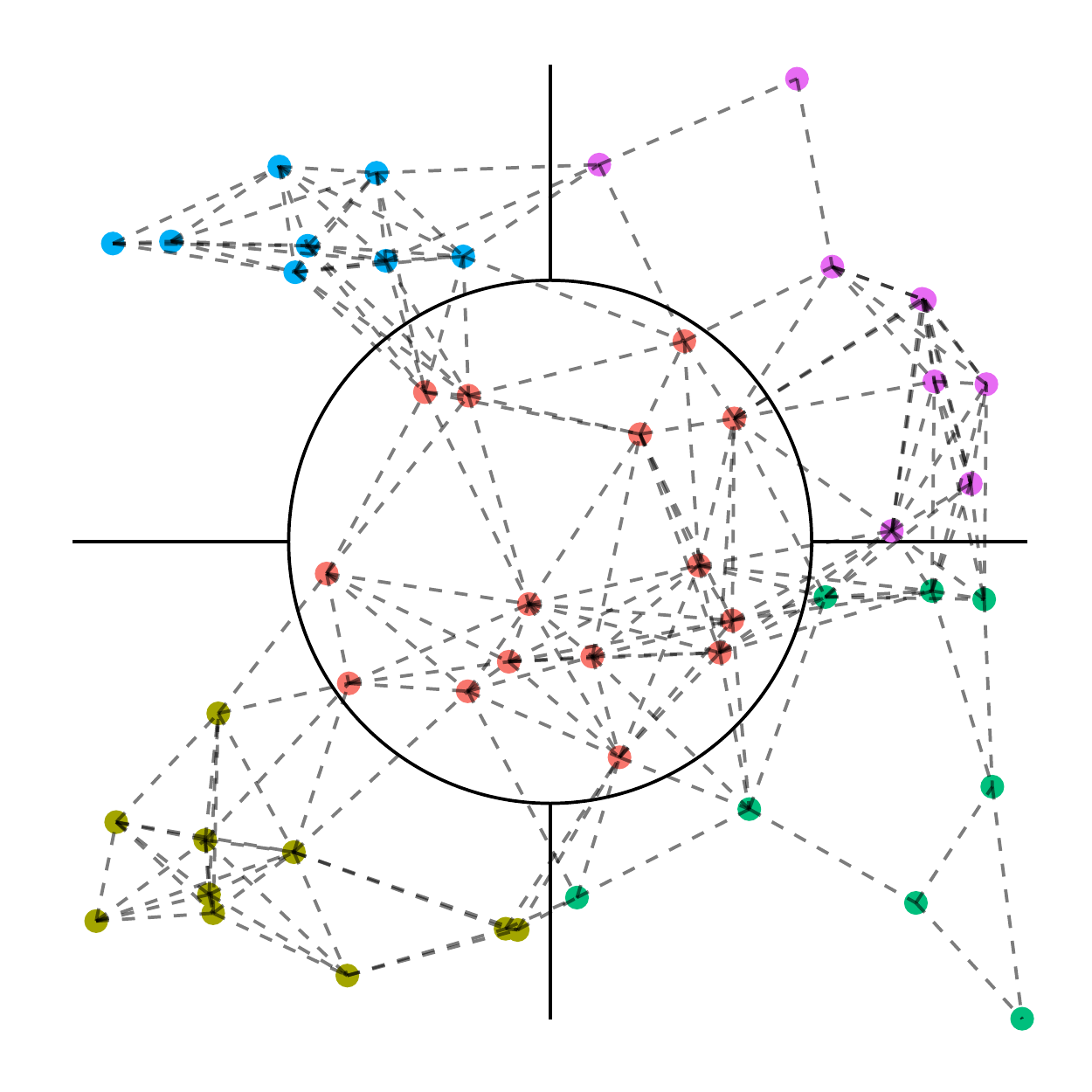} &
\includegraphics[width=.47\linewidth]{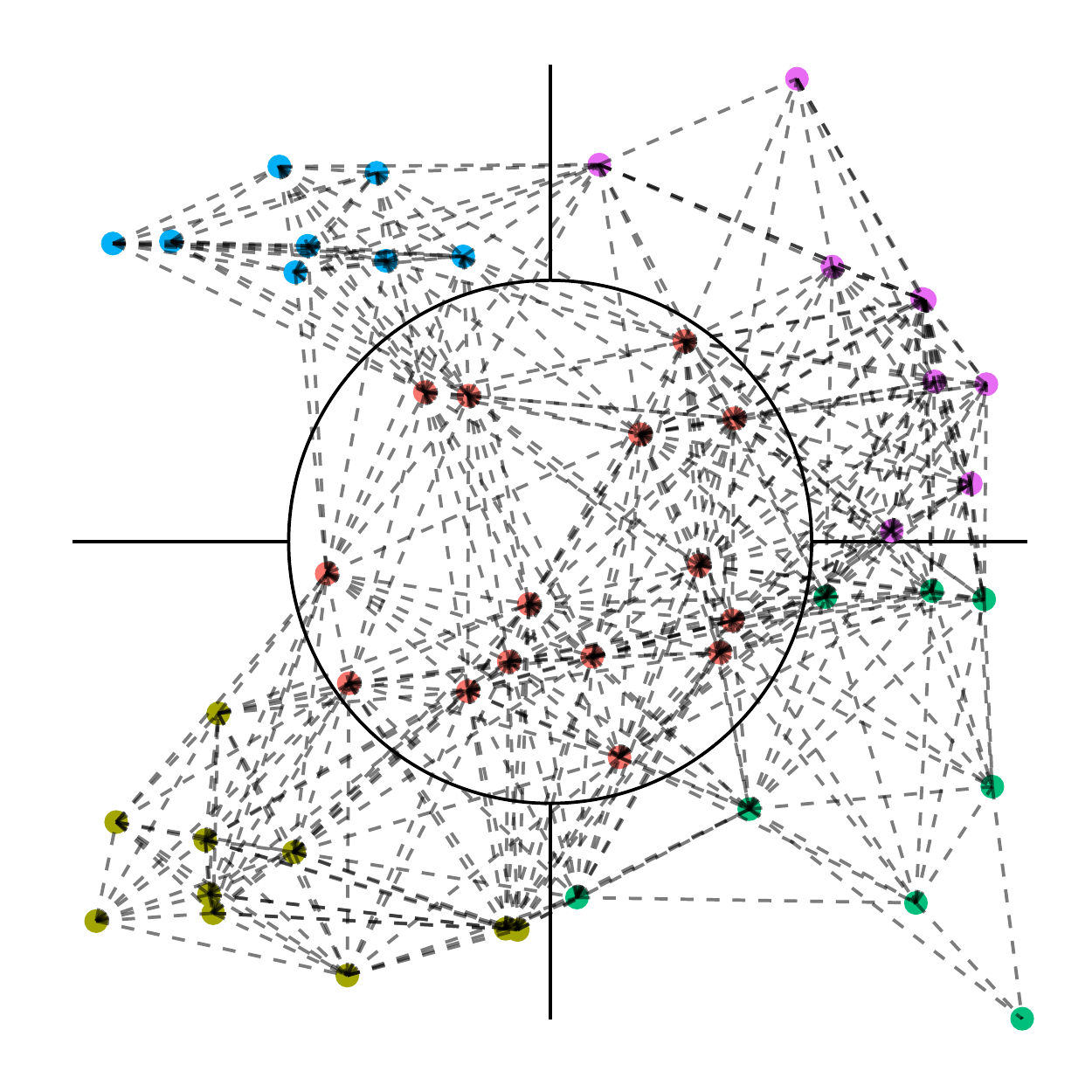} \\
{\small{(a) $r=0.5$.}} & {\small{(b) $r=0.75$.}} 
\end{tabular}
\vspace{-.1in}
\caption{A $50$-node network with two connection radiuses. 
Solid lines show the clusters and dash lines represent the edges in the network.}\label{Fig: Design}
\vspace{-.15in}
\end{wrapfigure}

\begin{figure}[!ht]
\centering
\begin{tabular}{@{}cccc@{}}
\includegraphics[width=0.23\linewidth]{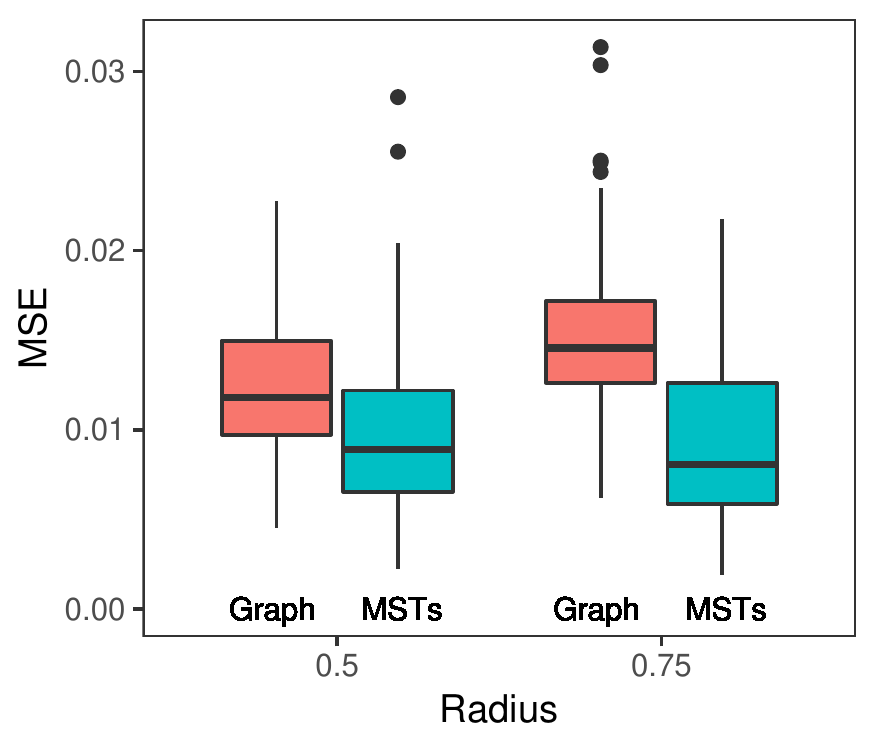}&
\includegraphics[width=0.23\linewidth]{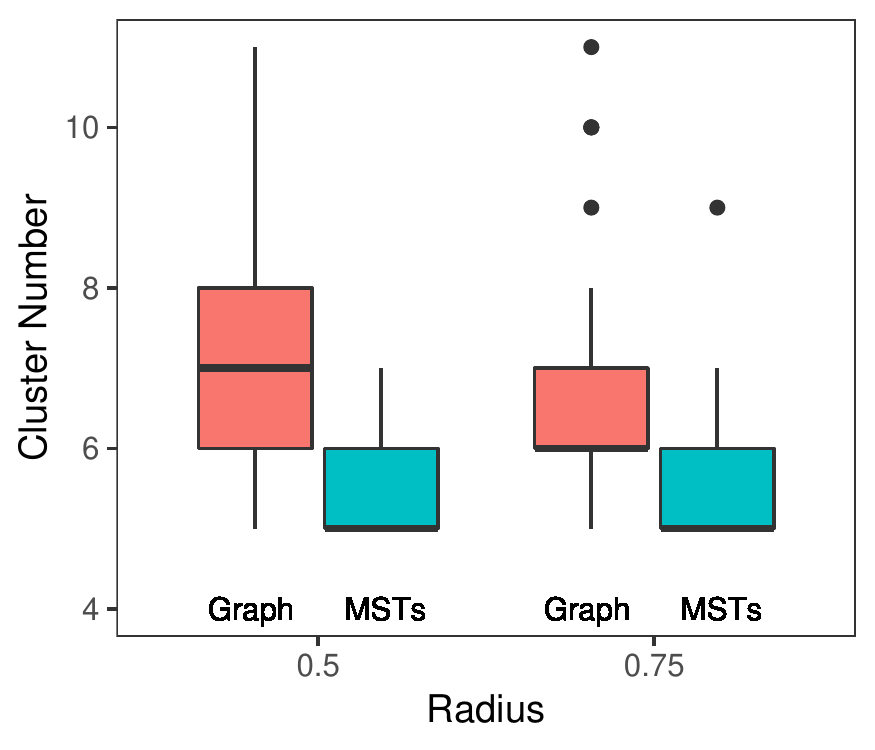}&
\includegraphics[width=0.23\linewidth]{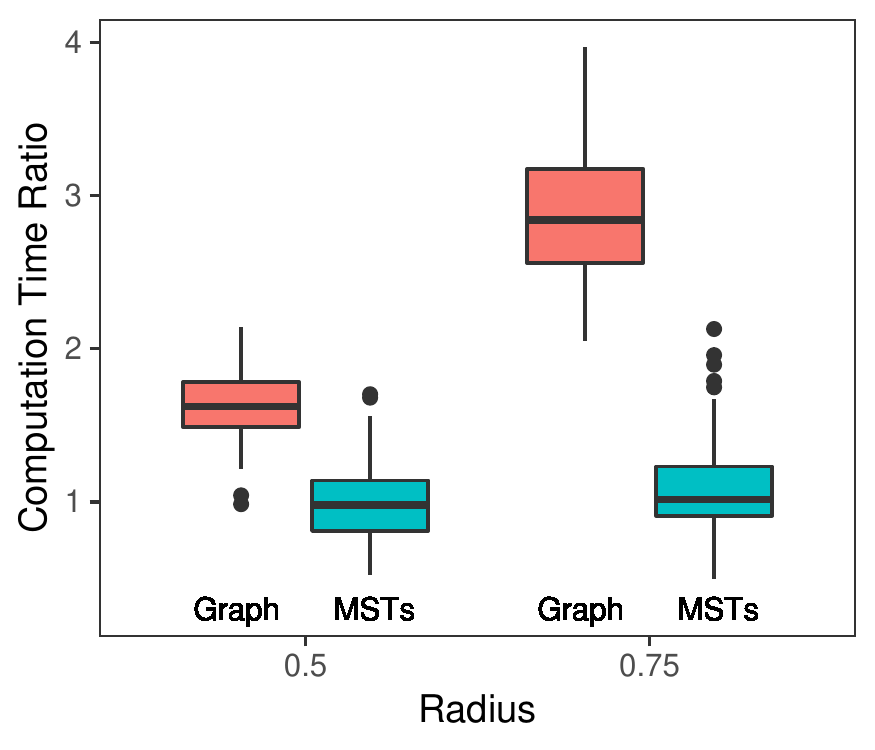}&
\includegraphics[width=0.23\linewidth]{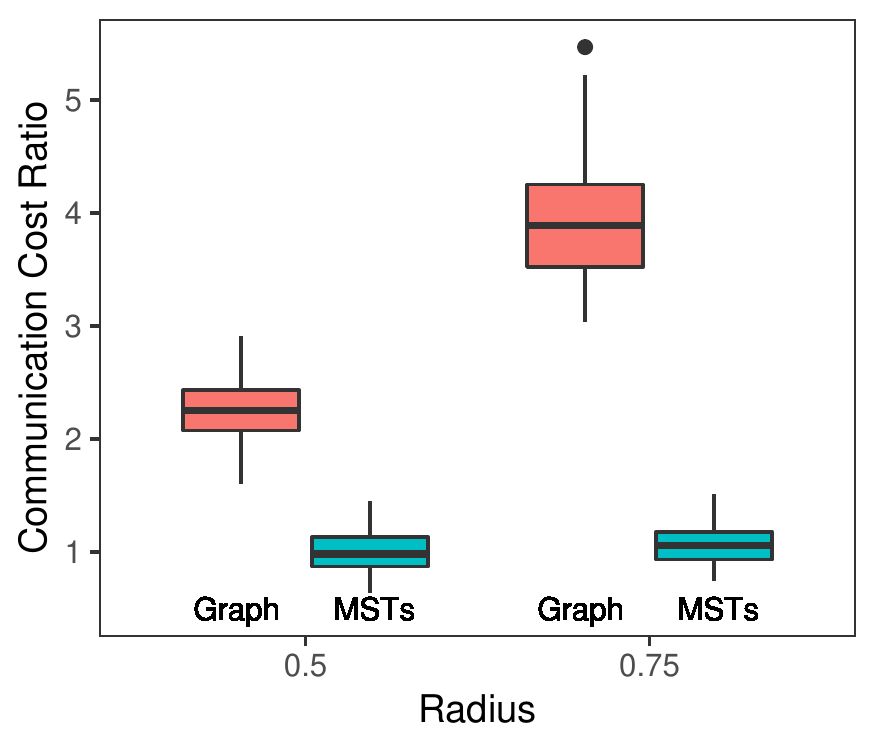}\\
\end{tabular}
\caption{The boxplots of MSEs of $\hat{\b}$, the estimated group numbers $\hat{S}$, the computation time ratio and the communication cost ratio of the graph $\ell_1$ regularization and MST$_s$ $\ell_1$ regularization.}\label{Fig: Simu3_compare}
\end{figure}

Due to space limitation, we only provide the numerical results of the impacts of the choices of regularization on accuracy and cost. 
More detailed numerical studies can be found in the supplementary materials.
%The computations are performed on a Windows computer with a 2.93 GHz Intel(R) Core(TM) i7 CPU processor and 16.0 GB memory. 
We compare our MST$_s$-based $\ell_1$ regularization ($\Theta(K)$ penalty terms) to the pairwise $\ell_{1}$ regularization ($O(K^2)$ penalty tems), which will be referred to as Graph-$\ell_1$ regularization in this section. 
Both models are solved by our proposed generalized ADMM algorithm distributively.
In the distributed algorithm, the nodes need to update the local $\b_i,$ $\vDelta_{i,l}$ and $\z_{i,l}$ in each iteration. 
%Note that the numbers of $\{\t_{i,l}\}_l$ and $\{\z_{i,l}\}_l$ are the same as those the penalty terms associated with node $i$. 
%Meanwhile, the nodes are required to send the local $\b_i$ to their neighbor nodes in the graph or MST.
Clearly, the amount of data being transmitted grows as the graph becomes denser.  
We simulate a $50$-node network and each node contains $50$ samples. 
We adjust the network denseness by changing the connection radius $r$. 
Two setting are compared: $r=0.50$ and $r=0.75$ (see Figure~\ref{Fig: Design} (a) -- (b)). 
We compare the accuracy and costs of the two models with $100$ simulations. 
The MSEs and the estimated cluster number $\hat{S}$ are used for measuring accuracy. 
%Here we only consider the synchronous algorithm for the computation time approximation. 
%Note that the network in Figure~\ref{Fig: Design} (b) with more edges takes longer time to exchange information. 
%Thus, the computational time of each iteration is proportional to the maximum node degree, and the total computational time is proportional to the summation of the running times of all information exchanges. 
%The communication cost is defined as the total amount of transmitted messages, which is proportional to the product of the iterations and the edges. 
We set the baseline to be the average computation time and the average communication cost for the MST$_s$ $\ell_1$ model under $r=0.50$. 
The boxplots for the accuracy, the computation time ratios, and the communication cost ratios are shown in Figure \ref{Fig: Simu3_compare}. 
We can see that our method outperforms in all aspects: 
Our method improves the MSE at least $21\%$, while reducing at least $38\%$ computation time and $55\%$ communication cost.

%By pruning redundant edges, our MST$_s$-based fused-lasso method enjoys both lower computation and communication costs, as well as higher estimation accuracy. 
%In contrast, for the Graph-$\ell_1$ model takes longer computational time and higher communication cost, as well as less accurate estimation. 
%Note that the estimated group number for the graph $\ell_1$ method is closer to the true group number with more complex network. 
%This is because the more edges in the graph, the less likely the graph breaks into the disjoint parts. 

\section{Conclusion} \label{Section: Conclusion}

In this work, we considered the problem of distributively learning the regression coefficient heterogeneity over networks.
We developed a new minimum spanning tree based adaptive fused-lasso model and a low-complexity distributed generalized ADMM algorithm to solve the problem.
We investigated the theoretical properties of both the model consistency and the algorithm convergence.
We showed that our model enjoys the oracle properties (i.e., selection consistency and asymptotic normality) and our distributed optimization algorithm has a linear convergence rate.
An interesting future topic is to generalize our framework to a more general class of regression problems including generalized linear model and semi-parameteric linear model.

\bibliography{reference.bib}

\end{document}